\documentclass{article} % For LaTeX2e
\usepackage{iclr2026_conference,times}

\usepackage{amsmath,amsfonts,bm}

\def\eqref#1{equation~\ref{#1}}
\def\1{\bm{1}}

\def\vs{{\bm{s}}}

\DeclareMathAlphabet{\mathsfit}{\encodingdefault}{\sfdefault}{m}{sl}
\SetMathAlphabet{\mathsfit}{bold}{\encodingdefault}{\sfdefault}{bx}{n}

\usepackage{hyperref}
\usepackage{url}
\usepackage{graphicx}
\usepackage{multirow}
\usepackage{wrapfig}
\usepackage{xspace}
\usepackage[dvipsnames]{xcolor}
\usepackage{amsmath}
\usepackage{pifont}
\usepackage{cleveref}
\usepackage{booktabs}
\usepackage[font=small]{caption}

\title{WAFT: Warping-Alone Field Transforms for Optical Flow}

\author{Yihan Wang \& Jia Deng \\
Department of Computer Science\\
Princeton University\\
\texttt{\{yw7685, jiadeng\}@princeton.edu} \\
}

\iclrfinalcopy % Uncomment for camera-ready version, but NOT for submission.
\begin{document}

% \newcolumntype{L}[1]{>{\raggedright\let\newline\\\arraybackslash\hspace{0pt}}m{#1}}
% \newcolumntype{C}[1]{>{\centering\let\newline\\\arraybackslash\hspace{0pt}}m{#1}}
% \newcolumntype{R}[1]{>{\raggedleft\let\newline\\\arraybackslash\hspace{0pt}}m{#1}} 
\newcommand{\xpar}[1]{\noindent\textbf{#1}\ \ }
\newcommand{\vpar}[1]{\vspace{3mm}\noindent\textbf{#1}\ \ }

\newcommand{\sect}[1]{Section~\ref{#1}}
\newcommand{\sects}[1]{Sections~\ref{#1}}
\newcommand{\eqn}[1]{Equation~\ref{#1}}
\newcommand{\eqns}[1]{Equations~\ref{#1}}
\newcommand{\fig}[1]{Figure~\ref{#1}}
\newcommand{\figs}[1]{Figures~\ref{#1}}
\newcommand{\tab}[1]{Table~\ref{#1}}
\newcommand{\tabs}[1]{Tables~\ref{#1}}
\newcommand{\x}{\mathbf{x}}
\newcommand{\y}{\mathbf{y}}
\newcommand{\fid}{Fr\'echet Inception Distance\xspace}
\newcommand{\lblfig}[1]{\label{fig:#1}}
\newcommand{\lblsec}[1]{\label{sec:#1}}
\newcommand{\lbleq}[1]{\label{eq:#1}}
\newcommand{\lbltbl}[1]{\label{tbl:#1}}
\newcommand{\lblalg}[1]{\label{alg:#1}}
\newcommand{\lblline}[1]{\label{line:#1}}

\newcommand{\ignorethis}[1]{}
\newcommand{\norm}[1]{\lVert#1\rVert_1}
\newcommand{\fcseven}{$\mbox{fc}_7$}

\newsavebox\CBox
\def\textBF#1{\sbox\CBox{#1}\resizebox{\wd\CBox}{\ht\CBox}{\textbf{#1}}}
\renewcommand*{\thefootnote}{\fnsymbol{footnote}}

\def\naive{na\"{\i}ve\xspace}
\def\Naive{Na\"{\i}ve\xspace}

\makeatletter
\DeclareRobustCommand\onedot{\futurelet\@let@token\@onedot}
\def\@onedot{\ifx\@let@token.\else.\null\fi\xspace}

\def\iid{\emph{i.i.d}\onedot}
\def\eg{\emph{e.g}\onedot} \def\Eg{\emph{E.g}\onedot}
\def\ie{\emph{i.e}\onedot} \def\Ie{\emph{I.e}\onedot}
\def\cf{\emph{c.f}\onedot} \def\Cf{\emph{C.f}\onedot}
\def\etc{\emph{etc}\onedot} \def\vs{\emph{vs}\onedot}
\def\wrt{w.r.t\onedot} \def\dof{d.o.f\onedot}
\def\etal{\emph{et al}\onedot}
\def\vs{\textbf{\emph{vs}\onedot}}
\makeatother

\definecolor{citecolor}{RGB}{34,139,34}
\definecolor{mydarkblue}{rgb}{0,0.08,1}
\definecolor{mydarkgreen}{rgb}{0.02,0.6,0.02}
\definecolor{mydarkred}{rgb}{0.8,0.02,0.02}
\definecolor{mydarkorange}{rgb}{0.40,0.2,0.02}
\definecolor{mypurple}{RGB}{111,0,255}
\definecolor{myred}{rgb}{1.0,0.0,0.0}
\definecolor{mygold}{rgb}{0.75,0.6,0.12}
\definecolor{mydarkgray}{rgb}{0.66, 0.66, 0.66}

\newcommand\scalemath[2]{\scalebox{#1}{\mbox{\ensuremath{\displaystyle #2}}}}
\newcommand{\supplement}[1]{\color{blue}{#1}}
\newcommand{\myparagraph}[1]{\paragraph{#1}}

\def\multirowcenter{-0.5\dimexpr \aboverulesep + \belowrulesep + \cmidrulewidth}
\renewcommand{\thefootnote}{\number\value{footnote}}

\newcommand\blfootnotetext[1]{%
  \begingroup
  \renewcommand\thefootnote{}\footnotetext{#1}%
  %\addtocounter{footnote}{-1}%
  \endgroup
}

\maketitle

\begin{abstract}
We introduce Warping-Alone Field Transforms (WAFT), a simple and effective method for optical flow. WAFT is similar to RAFT but replaces cost volume with high-resolution warping, achieving better accuracy with lower memory cost. This design challenges the conventional wisdom that constructing cost volumes is necessary for strong performance. WAFT is a simple and flexible meta-architecture with minimal inductive biases and reliance on custom designs. Compared with existing methods, WAFT ranks 1st on Spring, Sintel, and KITTI benchmarks, achieves the best zero-shot generalization on KITTI, while being $1.3-4.1\times$ faster than existing methods that have competitive accuracy (e.g., $1.3\times$ than Flowformer++, $4.1\times$ than CCMR+). Code and model weights are available at \href{https://github.com/princeton-vl/WAFT}{https://github.com/princeton-vl/WAFT}.
\end{abstract}

\section{Introduction}

Optical flow is a fundamental low-level vision task that estimates per-pixel 2D motion between video frames. It has many downstream applications, including 3D reconstruction and synthesis~\citep{ma2022multiview, zuo2022view}, action recognition~\citep{sun2018optical, piergiovanni2019representation, zhao2020improved}, frame interpolation~\citep{xu2019quadratic, liu2020video, huang2020rife}, and autonomous driving~\citep{geiger2013vision, menze2015object, janai2020computer}.

Cost volumes~\citep{sun2018pwc, ilg2017flownet} with iterative updates~\citep{teed2020raft, wang2024sea} has become a standard design in most state-of-the-art methods~\citep{sun2018pwc, dosovitskiy2015flownet, xu2017accurate, teed2020raft, huang2022flowformer,wang2024sea, morimitsu2025dpflow}, especially when both accuracy and efficiency are taken into account. Previous work~\citep{sun2018pwc, teed2020raft} regards cost volumes as a more effective representation than image features, as it explicitly models the visual similarity between pixels. 

However, constructing cost volumes is expensive in both time and memory~\citep{zhao2024hybrid, xu2023memory}. The cost increases quadratically with the radius of the neighborhood. As a result, cost volumes are often constructed from low resolution features, limiting the ability of the model to handle high-resolution input images.

In this paper, we challenge the conventional wisdom that cost volume is necessary for strong performance with high efficiency, and introduce Warping-Alone Field Transforms (WAFT), a simplified design that replaces cost volumes with warping and achieves state-of-the-art accuracy across various benchmarks with high efficiency.

For each pixel in frame 1, instead of computing its similarities against many pixels in frame 2, warping simply fetches the feature vector of the corresponding pixel given by the current flow estimate; this enables memory-efficient high-resolution processing and leads to better accuracy.

The design of WAFT is simple, with flow-specific designs kept to the minimum. WAFT consists of an input encoder that extracts features from individual input frames and a recurrent update module that iteratively updates flow. Compared to other RAFT-like architectures, WAFT is much simplified because it does not use cost volumes and has removed the context encoder that provides extra features for the update module. WAFT is designed to function as a meta-architecture for optical flow in the sense that the individual components including input encoder and the update unit do not require custom designs and can use existing off-the-shelf (pretrained) architectures. In our experiments, we evaluate different choices (such as ResNet~\citep{he2016deep} and DPT~\citep{ranftl2021vision}) that have different accuracy-efficiency trade-offs.  

WAFT achieves state-of-the-art performance across various benchmarks with high efficiency and a simple design. Using a Twins~\citep{chu2021twins} backbone pre-trained only on ImageNet, WAFT ranks first on Spring, second on KITTI, and is competitive on Sintel. It also achieves the best zero-shot cross-dataset generalization on KITTI. Using a stronger backbone, depth-pretrained DAv2~\citep{yang2024depthv2}, WAFT outperforms existing methods on all public benchmarks. We achieve this with standard network architectures for the sub-modules ~\citep{dosovitskiy2020image, he2016deep, ranftl2021vision}, removing custom designs typically needed in prior work, while being $1.3-4.1\times$ faster than existing methods that have competitive accuracy (e.g., $1.3\times$ than Flowformer++~\cite{shi2023flowformer++}, $4.1\times$ than CCMR+~\cite{jahedi2024ccmr}).

Our main contributions are two-fold: (1) we challenge the conventional wisdom that cost volume is a key component for achieving state-of-the-art accuracy and efficiency for optical flow; (2) we introduce WAFT, a warping-based meta-architecture that is simpler and achieves state-of-the-art accuracy with high efficiency. 

\begin{figure}[t]
    \centering
    \includegraphics[width=0.8\linewidth]{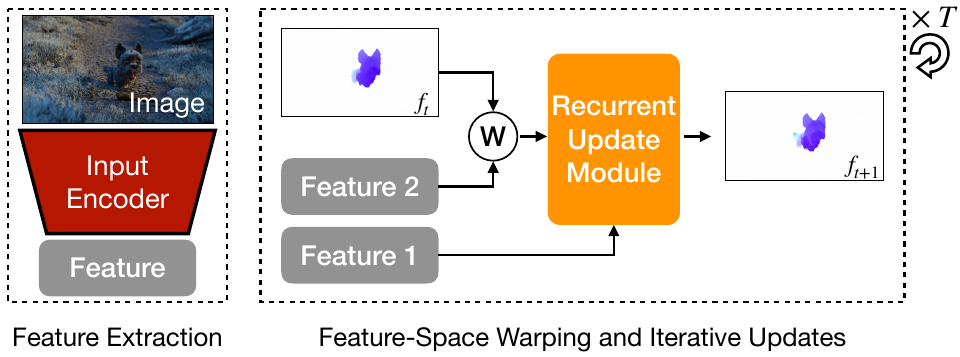}
    \caption{The meta-architecture of WAFT consists of an input encoder and a recurrent update module. We first extract image features from the input encoder, and then use these features to iteratively update the flow estimate for $T$ steps. At each step, we perform feature indexing through a lightweight backward warping on the feature of frame 2, removing the dependency on expensive cost volume used by previous work. }
    \label{fig:arch}
    \vspace{-12pt}
\end{figure}

\section{Related Work}

\myparagraph{Estimating Optical Flow} Traditional methods treated optical flow as a global optimization problem that maximizes visual similarity between corresponding pixels~\citep{horn1981determining, zach2007duality, chen2016full, brox2004high}. These methods apply coarse-to-fine warping~\citep{brox2004high, black1996robust, memin1998multigrid}, a strategy theoretically justified by Brox~\etal~\citep{brox2004high}, to solve this optimization. 

Today, this field is dominated by deep learning methods~\citep{ilg2017flownet, dosovitskiy2015flownet, sun2018pwc, zhao2020maskflownet, hui2018liteflownet, teed2020raft, sui2022craft, sun2022skflow, deng2023explicit, huang2022flowformer, shi2023flowformer++, weinzaepfel2022croco, weinzaepfel2023croco, xu2022gmflow, xu2023unifying, leroy2023win, saxena2024surprising, jahedi2024ccmr, jahedi2023ms, luo2022kpa, zheng2022dip, zhao2022global, luo2023gaflow, jung2023anyflow, luo2024flowdiffuser, zhou2024samflow, morimitsu2025dpflow}, which can be categorized into two paradigms: direct or iterative. Direct methods~\citep{dosovitskiy2015flownet, weinzaepfel2022croco, weinzaepfel2023croco, saxena2024surprising, leroy2023win, xu2022gmflow} treat flow estimation as a standard dense prediction task (\eg monocular depth estimation) and directly regress the dense flow field from large-scale pre-trained models. Iterative methods~\citep{teed2020raft, wang2024sea, luo2024flowdiffuser, morimitsu2025dpflow, zhou2024samflow, sun2018pwc, huang2022flowformer} align more closely with traditional warping-based approaches, refining the flow predictions progressively. Most state-of-the-art methods~\citep{teed2020raft, wang2024sea, huang2022flowformer, shi2023flowformer++, luo2024flowdiffuser, morimitsu2025dpflow} follow the iterative paradigm due to its significantly higher efficiency than the direct ones. 

Cost volumes~\citep{sun2018pwc, dosovitskiy2015flownet} have been regarded as a standard design in iterative methods~\citep{teed2020raft, wang2024sea, morimitsu2025dpflow, huang2022flowformer, shi2023flowformer++, xu2023unifying, luo2024flowdiffuser}. Prior work~\citep{teed2020raft, sun2018pwc} empirically shows the effectiveness of cost volumes in handling large displacements. Many iterative methods~\citep{teed2020raft, wang2024sea, morimitsu2025dpflow} adopt partial cost volumes~\citep{sun2018pwc} to avoid the quadratic computational complexity of full 4D cost volumes; they restrict the search range of each pixel in frame 1 to the neighborhood of its corresponding pixel in frame 2. However, they still suffer from the high memory consumption inherent in cost volumes~\citep{xu2023memory, zhao2024hybrid}.

WAFT is a warping-based iterative method. We achieve state-of-the-art performance across various benchmarks without constructing cost volumes, challenging the conventional wisdom established by previous work~\citep{sun2018pwc, teed2020raft, huang2022flowformer}. Warping no longer suffers from high memory consumption inherent in cost volumes, which enables high-resolution indexing and therefore improves accuracy.

\myparagraph{Vision Transformers} Vision transformers~\citep{dosovitskiy2020image} have achieved significant progress across a wide range of visual tasks~\citep{yang2024depthv2, kirillov2023segment, oquab2023dinov2, he2022masked, rombach2022high}. In the context of optical flow, most direct methods~\citep{weinzaepfel2022croco, weinzaepfel2023croco, saxena2024surprising} regress flow from a large-scale pre-trained vision transformer with a lightweight flow head. Many iterative methods~\citep{huang2022flowformer, shi2023flowformer++, luo2024flowdiffuser, zhou2024samflow} design task-specific transformer blocks to process cost volumes.

WAFT adopts similar designs to DPT~\citep{ranftl2021vision} in its recurrent update module, which implicitly handles large displacements in optical flow through the transformer architecture. We empirically show that this is crucial to make warping work. WAFT can also benefit from large-scale pre-trained transformers like existing methods~\citep{saxena2024surprising, weinzaepfel2022croco, weinzaepfel2023croco, zhou2024samflow}, with minimal additional flow-specific designs.

\section{Background}

In this section, we first review current cost-volume-based iterative methods and discuss the drawbacks of cost volumes. Then we introduce warping and compare it to cost volumes.

\begin{figure}[t]
    \centering
    \includegraphics[width=0.7\linewidth]{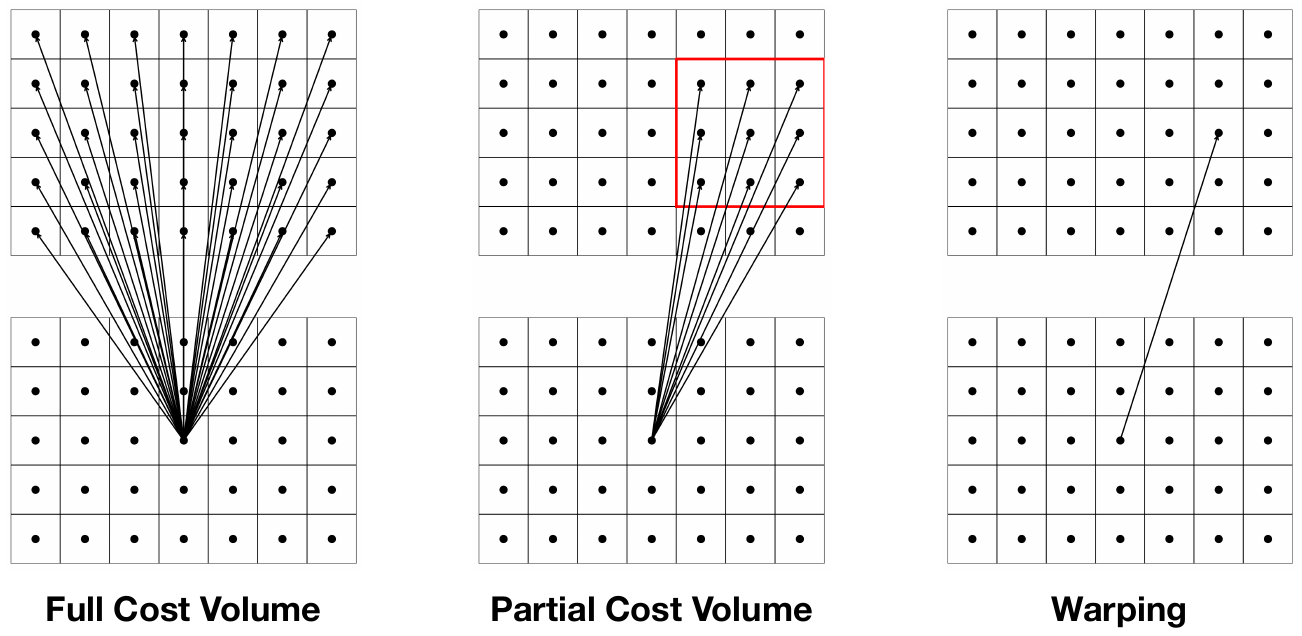}
    \caption{For each pixel, the full cost volume calculates its visual similarity to all pixels in the other frame through correlation. The partial cost volume restricts the search range to the neighborhood of the corresponding pixel, marked by a red box. Compared with them, warping only uses the information from the corresponding pixel, offering better time and memory efficiency. This efficiency enables high-resolution processing, which leads to improved accuracy.}
    \label{fig:diff}
    \vspace{-12pt}
\end{figure}

\subsection{Iterative Methods with Cost Volume}
\label{sec:iter-cost}

Given two adjacent RGB frames, optical flow predicts pixel-wise 2D motion between adjacent frames. Current iterative methods~\citep{teed2020raft, huang2022flowformer, wang2024sea} consist of two parts: (1) input encoders that extract dense image features at low resolution, and (2) a recurrent update module that iteratively refines the flow estimate.

Denoting the two frames as $I_1, I_2\in\mathbb{R}^{H\times W \times 3}$, the input encoder $F$ maps $I_1$, $I_2$ to low-resolution dense features $F(I_1), F(I_2)\in \mathbb{R}^{h\times w\times d}$, respectively. A cost volume $V\in\mathbb{R}^{h\times w\times h\times w}$ is built on these features, which explicitly models the correlation between pixels ($p\in I_1$, $p'\in I_2$):
$$V_{p, p'} = F(I_1)_{p}\cdot F(I_2)_{p'}$$ where $\cdot$ represents the dot product of two vectors. At each step, the recurrent update module indexes into the cost volume using the current flow estimate and predicts the residual flow update. Several methods~\citep{huang2022flowformer, shi2023flowformer++} directly process $V$ for better performance, but tend to be more costly.

Partial cost volume~\citep{sun2018pwc} is introduced to reduce the cost by avoiding the construction of a full 4D cost volume. Given the current flow estimate $f_{cur}\in\mathbb{R}^{h\times w\times 2}$ and a pre-defined look-up radius $r$, partial cost volume $V_{par}: \mathbb{R}^{h\times w\times 2}\to\mathbb{R}^{h\times w\times (r^2)}$ implements an on-the-fly partial construction by restricting the indexing range of a pixel $p\in I_1$ to the neighborhood of its corresponding pixel $p+(f_{cur})_p\in I_2$, formulated as: $$V_{par}(f_{cur};r)_{p} = \text{concat}(\{V_{p, p'}|\forall p'\in I_{2}, \text{s.t.} \|p+(f_{cur})_{p}-p'\|_{\infty}\le r\|)$$

where the operator ``concat'' concatenates all values inside the set into a vector. In practice, partial cost volumes are usually constructed at multiple scales~\citep{teed2020raft, wang2024sea} to improve the prediction of large displacements. Current methods also introduce context encoder~\citep{sun2018pwc, teed2020raft, huang2022flowformer, wang2024sea} to enhance the effectiveness of iterative refinement.

\subsection{Drawbacks of Cost Volumes}
\label{sec:drawbacks-cost-volume}

\begin{figure}[t]
    \centering
    \includegraphics[width=0.8\linewidth]{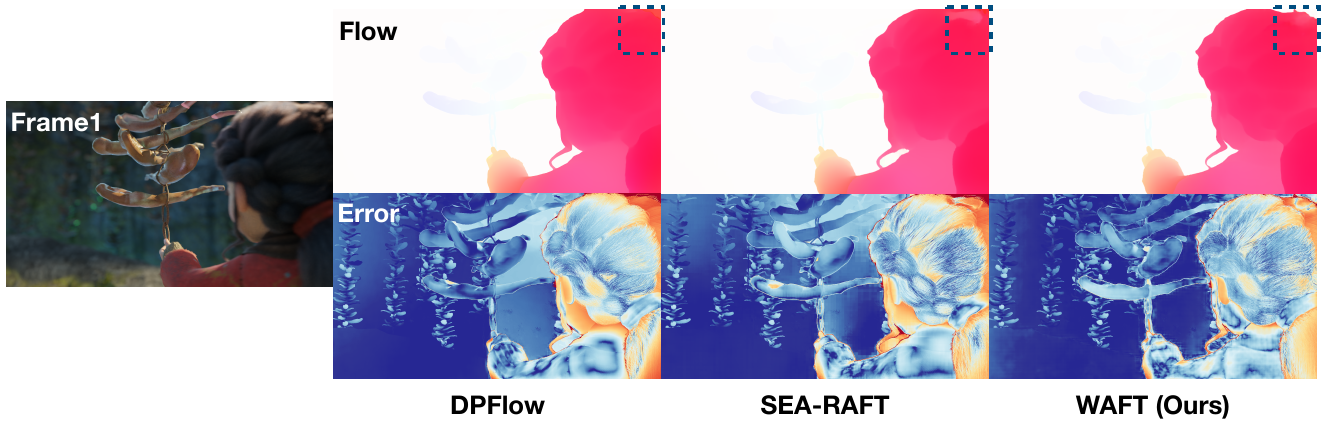}
    \caption{Visualizations of different methods on Spring~\citep{mehl2023spring}. WAFT, benefiting from high-resolution indexing, obtains sharper boundaries and lower errors than low-resolution approaches.}
    \label{fig:high-res}
    \vspace{-15pt}
\end{figure}

\begin{wraptable}[11]{r}{0.5\textwidth}
\centering
\vspace{-11pt}
\resizebox{0.5\textwidth}{!}{
\begin{tabular}{lccc}
\toprule
\multirow{2}{*}[\multirowcenter]{Method} & \multicolumn{3}{c}{Training Memory Cost (GiB)}\\
\cmidrule(l{0.5ex}r{0.5ex}){2-4}
& 1/8 Reso. & 1/4 Reso. & 1/2 Reso. \\
\midrule
SEA-RAFT & 14.1 & 25.8 & OOM \\
Flowformer & 26.1 & - & - \\
CCMR+ & 36.0 & - & - \\
WAFT-Twins-a2 & 7.0 & 7.6 & 9.2 \\
\bottomrule
\end{tabular}
}
\caption{We profile the training memory cost with batch size 1 on an RTX A6000. Our warping method significantly reduces the cost.}
\label{tab:memory}
\end{wraptable}

\paragraph{High Memory Cost} The main drawback of cost volume is its high memory consumption. Full or partial cost volume at high resolution are very expensive and often infeasible~\citep{zhao2024hybrid, xu2023memory}. Therefore, most iterative methods build the cost volume and index into it at 1/8 resolution. To further demonstrate the problem, we implement several variants of SEA-RAFT~\citep{wang2024sea} that build partial cost volumes at different resolutions. We set the base channel dimension as 32, 64, and 128 for the 1/2, 1/4, and 1/8 resolution variants, respectively, to make their computational cost similar (around 350GMACs). As shown in Table~\ref{tab:memory}, the partial cost volume with look-up radius $r=4$ in SEA-RAFT runs out of memory at 1/2 resolution. WAFT removes the reliance on cost volume, and therefore consumes significantly lower memory than existing methods~\citep{wang2024sea, huang2022flowformer, jahedi2024ccmr}.

\paragraph{Error from Low Resolution Indexing} Since cost volumes are restricted to low resolution, the predicted flow field must be downsampled for cost volume look-up, which inevitably introduces errors. As illustrated in Figure~\ref{fig:high-res} using an example from the Spring benchmark~\citep{mehl2023spring}, existing methods struggle to produce clear boundaries, particularly noticeable in the top-right corner. In contrast, our method benefits from high-resolution look-up of feature vectors and obtains sharper boundary predictions in these challenging regions. We also show the quantitative performance gain from high-resolution indexing in Table~\ref{tab:ablations}. 

\subsection{Warping vs. Cost Volume}
\label{sec:warp}

Warping was widely used in both classical and early deep learning approaches~\citep{memin1998multigrid, brox2004high, ilg2017flownet, ranjan2017optical}. However, most recent iterative methods~\citep{wang2024sea, teed2020raft, huang2022flowformer} have either replaced warping with cost volumes or used both in combination, since cost volumes have been shown to remarkably improve the performance~\citep{sun2018pwc}. In this section, we analyze the similarities and differences between warping and cost volume (see Figure~\ref{fig:diff}), and argue that with appropriate designs, warping-based methods can achieve performance on par with cost-volume-based methods.

The overlap of warping and cost volumes lies in their use of the current flow prediction to index into feature maps, which is closely related to optimization~\citep{brox2004high}. In cost-volume-based iterative methods, the current flow estimate $f_{cur}$ is used to define partial cost volume $V_{par}$ (see Section~\ref{sec:iter-cost}). It is also used to define the warped feature map $\texttt{Warp}(f_{\text{cur}})\in\mathbb{R}^{h\times w\times d}$, where the feature vector of pixel $p\in I_{1}$ is indexed from the feature map of frame 2, formulated as: $$\texttt{Warp}(f_{\text{cur}})_{p}=F(I_2)_{p+(f_{\text{cur}})_p}$$

Compared to cost volumes, for each pixel in frame 1, warping does not calculate its visual similarity to multiple pixels in frame 2, making it no longer able to explicitly model large displacements. However, we can implicitly handle this long-range dependence through the attention mechanism in vision transformers~\citep{dosovitskiy2020image, ranftl2021vision}, which, as we will demonstrate, is crucial to make warping work well (see Section~\ref{sec:ablation}). The high memory efficiency of warping also enables high-resolution indexing, leading to improved accuracy.

\section{Method}

\begin{figure}[t]
    \centering
    \includegraphics[width=0.7\linewidth]{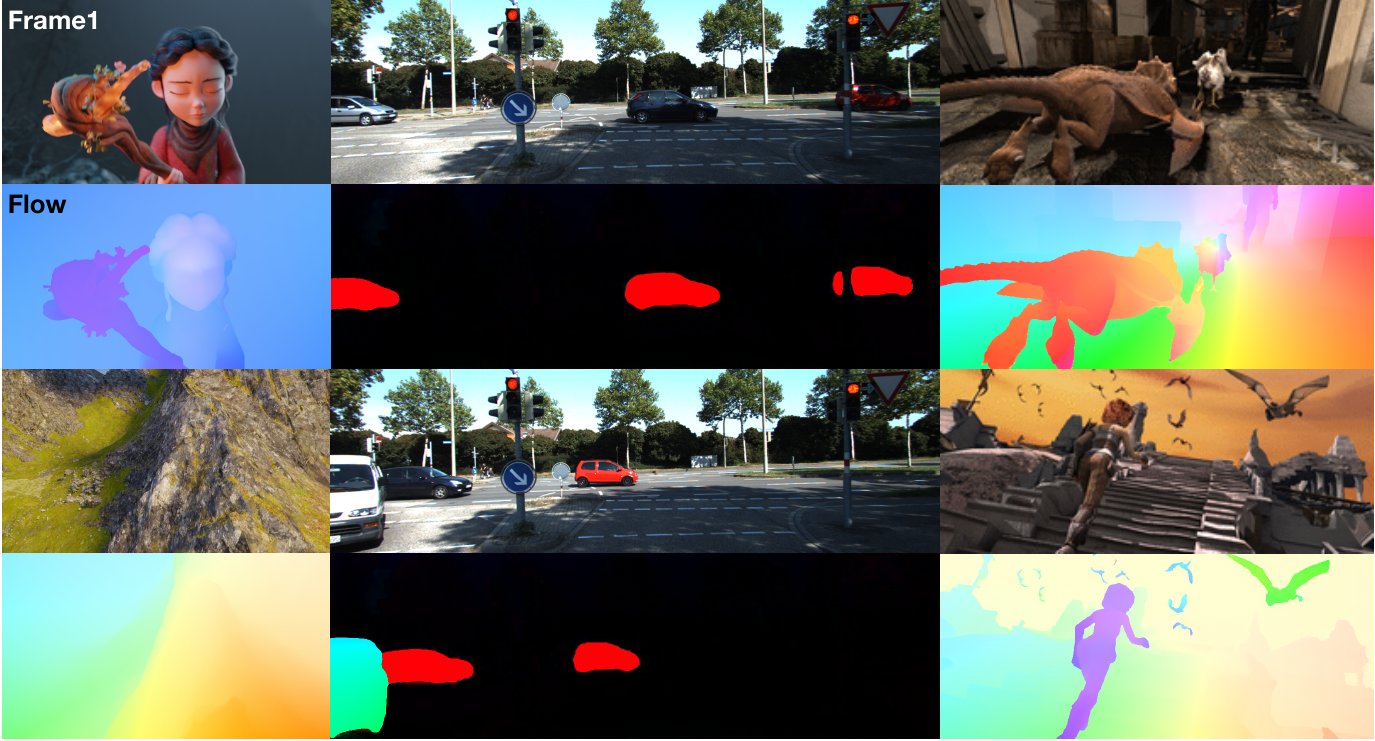}
    \caption{Visualizations on Spring, KITTI, and Sintel public benchmarks (from left to right).}
    \label{fig:vis}
    \vspace{-12pt}
\end{figure}

In this section, we describe WAFT, our warping-based iterative method, shown in Figure~\ref{fig:arch}. Its design can be understood as a simple meta-architecture that integrates an input encoder and a recurrent update module. We will also discuss the advantages of our design, especially on its strong performance and the simplifications over past designs.

\myparagraph{Input Encoder} We develop two ways to adapt large-scale pre-trained models. Adaptation 1 (a1) is our initial design specific to DAv2. We freeze the entire DAv2, incorporate features from its DPT head, and further refine these features using a ResNet18. Adaptation 2 (a2) is our improved design that works better and supports more backbones, where we only freeze the ViT/CNN backbones. We make the DPT head trainable and side-tune the features with 3 ResNet blocks. For completeness, we report the results of both.

\myparagraph{Recurrent Update Module} Similar to existing iterative methods~\citep{teed2020raft, wang2024sea, huang2022flowformer}, our recurrent update module $R$ iteratively predicts the residual flow updates. At step $t$, we concatenate $F(I_1)$ (feature of frame 1), $\texttt{Warp}(f_{\text{cur}})$ (warped feature of frame 2, see Section~\ref{sec:warp}), and the current hidden state $\text{Hidden}_{t}\in\mathbb{R}^{h\times w\times d}$ as input. We use a slightly modified DPT~\citep{ranftl2021vision} as the architecture of the module.

\myparagraph{Prediction Head \& Loss} We adopt the Mixture-of-Laplace (MoL) loss used in SEA-RAFT~\citep{wang2024sea}. At step $t$, the hidden state $\text{Hidden}_{t}$ is used to predict the MoL parameters $M\in\mathbb{R}^{h\times w\times 6}$. They are upsampled to the original image resolution through convex upsampling~\citep{teed2020raft, wang2024sea}.

\myparagraph{Simplifications over Existing Iterative Methods} We replace cost volumes, which are standard in existing iterative methods~\citep{sun2018pwc, teed2020raft, huang2022flowformer, wang2024sea}, with high-resolution warping, which is more memory-efficient. In addition, we have removed the context encoder~\citep{sun2018pwc}, another flow-specific design standard in existing iterative methods. 

A direct benefit from our simplified design is that we can load pre-trained weights for standard architectures such as ViT~\citep{dosovitskiy2020image}, which can improve generalization as our experiments will show Section~\ref{sec:ablation}. 

The simplicity of our meta-architecture also enables more apples-to-apples comparisons between direct methods and iterative methods. Existing direct methods~\citep{weinzaepfel2022croco, weinzaepfel2023croco, saxena2024surprising} share an input format similar to that of the first iteration of WAFT, making them more directly compatible. We will empirically show the effectiveness and necessity of iterative indexing within our meta-architecture in Section~\ref{sec:ablation}.

\section{Experiments}

We report results on Sintel~\citep{butler2012naturalistic}, KITTI~\citep{geiger2013vision}, and Spring~\citep{mehl2023spring}. Following existing work~\citep{teed2020raft, huang2022flowformer, wang2024sea, morimitsu2025dpflow}, for training, we use FlythingChairs~\citep{dosovitskiy2015flownet}, FlyingThings~\citep{mayer2016large}, HD1K~\citep{kondermann2016hci}, Sintel~\citep{butler2012naturalistic}, KITTI~\citep{geiger2013vision}, Spring~\citep{mehl2023spring}, and TartanAir~\citep{wang2020tartanair}. We use the widely adopted metrics: endpoint-error (EPE), 1-pixel outlier rate (1px), percentage of flow outliers (Fl), and weighted area under the curve (WAUC). Definitions can be found in~\citep{richter2017playing, mehl2023spring, geiger2013vision, morimitsu2025dpflow}.

\subsection{Architecture Details} 

\setlength\tabcolsep{.4em}
\begin{table*}[t]
    \centering
    \resizebox{1.0\linewidth}{!}{
    \begin{tabular}{llcccccc}
    \toprule
    \multirow{2}{*}[\multirowcenter]{Type} & \multirow{2}{*}[\multirowcenter]{Method} & \multicolumn{2}{c}{Sintel} & \multicolumn{2}{c}{KITTI} & \multicolumn{2}{c}{Inference Cost}\\
    \cmidrule(l{0.5ex}r{0.5ex}){3-4}\cmidrule(l{0.5ex}r{0.5ex}){5-6}\cmidrule(l{0.5ex}r{0.5ex}){7-8}
    & & Clean$\downarrow$ & Final$\downarrow$ & All$\downarrow$ & Non-Occ$\downarrow$ & \#MACs (G) & Latency (ms)\\ 
    \midrule
        \multirow{3}{*}{Direct} & GMFlow~\citep{xu2022gmflow} & 1.74 & 2.90 & 9.32 & 3.80 & 603 & 139\\
        & CroCoFlow~\citep{weinzaepfel2023croco} & 1.09 & 2.44 & 3.64 & 2.40 & 57343 & 6422 \\
        & DDVM~\citep{saxena2024surprising} &1.75 &2.48 & \textBF{3.26} & 2.24 & -& -\\
    \midrule
        \multirow{16}{*}{\shortstack{Iterative\\w/ Cost Volume}} & PWC-Net+~\citep{sun2019models} & 3.45 & 4.60 & 7.72 & 4.91 & 101 & 24\\
        & RAFT~\citep{teed2020raft} & 1.61 & 2.86 & 5.10 & 3.07 & 938 & 141\\
        & DIP~\citep{zheng2022dip} &1.44 & 2.83 & 4.21 & 2.43 & 3068 & 499\\
        & GMFlowNet~\citep{zhao2022global} & 1.39 & 2.65 & 4.79 & 2.75 & 1094 & 244\\
        & CRAFT~\citep{sui2022craft} & 1.45 & 2.42 & 4.79 & 3.02 & 2274 & 483\\
        & FlowFormer~\citep{huang2022flowformer} & 1.20 & 2.12 & 4.68 & 2.69 & 1715 & 336\\
        & GMFlow+~\citep{xu2023unifying} & 1.03 & 2.37 & 4.49 & 2.40 & 1177 & 250\\
        & RPKNet~\citep{morimitsu2024recurrent} & 1.31 & 2.65 & 4.64 & 2.71 & 137 & 183 \\ 
        & CCMR+~\citep{jahedi2024ccmr} & 1.07 & 2.10 & 3.86 & 2.07 & 12653 & 999 \\ 
        & MatchFlow(G)~\citep{dong2023rethinking} & 1.16 & 2.37 & 4.63 & 2.77 & 1669 & 291\\
        & Flowformer++\citep{shi2023flowformer++} & 1.07 & \textBF{1.94} & 4.52 & - & 1713 & 374\\ 
        & SEA-RAFT(L)~\citep{wang2024sea} &1.31 & 2.60 & 4.30 & - & 655 & 108\\
        & AnyFlow~\citep{jung2023anyflow} & 1.23 & 2.44 & 4.41 & 2.69 & - & - \\
        & FlowDiffuser~\citep{luo2024flowdiffuser} & 1.02 & 2.03 & 4.17 & 2.82 & 2466 & 599 \\
        & SAMFlow~\citep{zhou2024samflow} & 1.00 & 2.08 & 4.49 & - & 9717 & 1757\\
        & DPFlow~\citep{morimitsu2025dpflow} &  1.04 & 1.97 & 3.56 & 2.12 & 414 & 131 \\
    \midrule
        \multirow{6}{*}{\shortstack{Iterative\\ w/ Warping}} & SpyNet~\citep{ranjan2017optical} & 6.64 & 8.36 &  35.07 & 26.71 & 167 & 25 \\ 
        & FlowNet2~\citep{ilg2017flownet} & 4.16 & 5.74 & 10.41 & 6.94 & 230 & 75 \\
        & \textBF{WAFT-DAv2-a1} & 1.09 & 2.34 & 3.42 & \textBF{2.04} & 853 & 240 \\
        & \textBF{WAFT-Twins-a2} & 1.02 & 2.39 & 3.53 & 2.12 & 1020 & 290 \\
        & \textBF{WAFT-DAv2-a2} & \textBF{0.95} & 2.33 & 3.31 & \textBF{2.03} & 807 & 240 \\
        & \textBF{WAFT-DINOv3-a2} & \textBF{0.94} & 2.02 & 3.56 & 2.13 & 732 & 212 \\
    \bottomrule
    \end{tabular}
    }
    \caption{We report endpoint-error (EPE) on Sintel~\citep{butler2012naturalistic}, Fl on KITTI~\citep{geiger2013vision}, and highlight all SOTA performance. On KITTI, WAFT ranks first on non-occluded pixels and second on all pixels. It also achieves state-of-the-art performance on Sintel (clean). We measure the latency on an RTX3090 with batch size 1 and 540p input.}
    \label{tab:Sintel&KITTI}
    \vspace{-15pt}
\end{table*}

\setlength\tabcolsep{.4em}
\begin{wraptable}[14]{r}{0.5\textwidth}
\centering
\vspace{-12pt}
\resizebox{0.5\textwidth}{!}{
\begin{tabular}{lcccc}
\toprule
Method & 1px$\downarrow$ & EPE$\downarrow$ & Fl$\downarrow$ & WAUC$\uparrow$\\
\midrule
    FlowNet2~\citep{ilg2017flownet}$^*$ &  6.710 &  1.040 & 2.823 & 90.907 \\
    SpyNet~\citep{ranjan2017optical}$^*$ & 29.963 & 4.162 & 12.866 & 67.150\\
    PWC-Net~\citep{sun2018pwc}$^*$ &  82.27 & 2.288 & 4.889 & 45.670\\
    RAFT~\citep{teed2020raft}$^*$ &  6.790 & 1.476 & 3.198 & 90.920\\
    GMA~\citep{jiang2021learning}$^*$ &  7.074 & 0.914 & 3.079 & 90.722\\
    FlowFormer~\citep{huang2022flowformer}$^*$ & 6.510 & 0.723 & 2.384 & 91.679\\
    GMFlow~\citep{xu2022gmflow}$^*$ & 10.355 & 0.945 & 2.952 & 82.337\\
    RPKNet~\citep{morimitsu2024recurrent} & 4.809 & 0.657 & 1.756 & 92.638\\
    CroCoFlow~\citep{weinzaepfel2023croco} & 4.565 & 0.498 & 1.508 & 93.660\\
    Win-Win~\citep{leroy2023win} & 5.371 & 0.475 & 1.621 & 92.270\\
    SEA-RAFT(M)~\citep{wang2024sea} & 3.686 & 0.363 & 1.347 & 94.534\\
    DPFlow~\citep{morimitsu2025dpflow} & 3.442 & 0.340 & 1.311 & 94.980\\
    \textBF{WAFT-DAv2-a1-540p} & \textBF{3.418} & 0.340 & \textBF{1.280} & 94.663 \\
    \textBF{WAFT-DAv2-a1-1080p} & \textBF{3.347} & \textBF{0.337} & \textBF{1.222} & \textBF{95.189} \\
    \textBF{WAFT-Twins-a2} & \textBF{3.268} & \textBF{0.331} & \textBF{1.282} & 94.786 \\
    \textBF{WAFT-DAv2-a2} & \textBF{3.298} & \textBF{0.304} & \textBF{1.197} & \textBF{94.990} \\
    \textBF{WAFT-DINOv3-a2} & \textBF{3.182} & \textBF{0.325} & \textBF{1.246} & \textBF{95.051} \\
    \bottomrule
\end{tabular}
}
\caption{WAFT ranks 1st on Spring~\citep{mehl2023spring} on all metrics. We highlight all SOTA performance. $^{*}$ denotes the submissions from the Spring team.}
\label{tab:Spring}
\end{wraptable}

\myparagraph{Input Encoder} We use frozen ImageNet-pretrained Twins-SVT-Large~\citep{chu2021twins}, depth-pretrained DAv2-S~\citep{yang2024depthv2}, and unsupervised-pretrained DINOv3-ViT-S~\citep{simeoni2025dinov3} in input encoders.

\myparagraph{Recurrent Update Module} We use a modified DPT-Small~\citep{ranftl2021vision} as the recurrent update module. We concatenate the image features and use a $1\times 1$ conv to obtain the initial hidden state. Since the image features are already $2\times$ downsampled, we change the patch size to 8. We set the resolution of the positional embedding to $224\times 224$, and interpolate it for other resolutions. We use $T=5$ iterations in training and inference.

\subsection{Training Details}

\myparagraph{Benchmark Submissions} Following SEA-RAFT~\citep{wang2024sea}, we first pre-train our model on TartanAir~\citep{wang2020tartanair} for 300k steps, with a batch size of 32 and learning rate $4\times 10^{-4}$. We fine-tune our model on FlyingChairs~\citep{zhao2020maskflownet} with the same hyperparameters for 50k steps, and then fine-tune it on FlyingThings~\citep{mayer2016large} for 200k steps. For all submissions, we keep the batch size as 32 by default and reduce the learning rate to $10^{-4}$. For KITTI~\citep{geiger2013vision} submission, we fine-tune our model on KITTI(train) for 5k steps. For Sintel~\citep{butler2012naturalistic} submission, we follow previous work to fine-tune our model on the mixture of FlyingThings, HD1K~\citep{kondermann2016hci}, KITTI(train), and Sintel(train) for 200k steps. For Spring~\citep{mehl2023spring} submissions, we fine-tune our models on Spring(train) for 200k steps with a batch size of 32. We train an extra 1080p WAFT-DAv2-a1 model with a batch size of 8.

\myparagraph{Zero-Shot Evaluation} We first train our model on FlyingChairs for 50k steps, and then fine-tune it on FlyingThings for 50k steps. The batch size is set to 32, and the learning rate is set to $10^{-4}$.

\subsection{Benchmark Results} 

\myparagraph{Sintel \& KITTI} Results are shown in Table~\ref{tab:Sintel&KITTI}. Using a Twins backbone only pre-trained on ImageNet, WAFT ranks second on KITTI and is competitive on Sintel (clean). It outperforms prior cost-volume-based SOTA Flowformer++~\citep{shi2023flowformer++} in both accuracy and efficiency given the same backbone, demonstrating the strength of high-resolution warping. The performance can be further improved with stronger backbones~\citep{yang2024depthv2, simeoni2025dinov3}. Using a depth-pretrained DAv2~\citep{yang2024depthv2}, on KITTI~\citep{geiger2013vision}, WAFT achieves the best Fl on non-occluded pixels and the second best on all pixels. It also ranks first on Sintel (Clean)~\citep{butler2012naturalistic} and competitive on Sintel (Final). WAFT is 1.3-4.1$\times$ faster than existing methods that have competitive accuracy (e.g., 1.3$\times$ than Flowformer++~\cite{shi2023flowformer++}, 4.1$\times$ than CCMR+~\cite{jahedi2024ccmr}), demonstrating its high efficiency.

Note that there is an outlier sequence, `Ambush 1', which severely affects the average performance on Sintel (Final) as mentioned in~\cite{saxena2024surprising}. We show that WAFT outperforms Flowformer++ on Sintel (Final) with the same Twins backbone when `Ambush 1' is excluded. More details can be found in Table~\ref{tab:sintel-seq}. 

\myparagraph{Spring} Results are shown in Table~\ref{tab:Spring}. Following the downsample-upsample protocol (540p) of SEA-RAFT~\citep{wang2024sea}, WAFT outperforms existing methods on EPE and 1px with a Twins backbone only pre-trained on ImageNet. We also show that WAFT achieves the best performance on all metrics with a depth-pretrained DAv2 backbone. Benefiting from warping, WAFT can be trained at full resolution (1080p) to improve the performance further.

\myparagraph{Comparison with Existing Warping-based Methods} It appears that warping as a network operation has been largely abandoned by works in the last 8 years and the last time warping-based methods achieved top positions on the leaderboards were around 2017~\citep{ranjan2017optical, ilg2017flownet}. WAFT is significant in that it has revisited and revived an idea that has fallen out of favor. Compared to methods that do use warping~\citep{ranjan2017optical, ilg2017flownet}, WAFT reduces endpoint-error (EPE) by at least 64\% on Sintel and 70\% on Spring~\citep{mehl2023spring}, while also reducing Fl by at least 68\% on KITTI~\citep{geiger2013vision} and 57\% on Spring~\citep{mehl2023spring}. Besides, WAFT reduces 1px-outlier rate by 52\% on Spring~\citep{mehl2023spring}. 

\subsection{Zero-Shot Evaluation}

\begin{wraptable}[17]{r}{0.5\textwidth}
\centering
\vspace{-35pt}
\resizebox{0.5\textwidth}{!}{
\begin{tabular}{lcccc}
\toprule
\multirow{2}{*}[\multirowcenter]{Method} & \multicolumn{2}{c}{Sintel (train)} & \multicolumn{2}{c}{KITTI (train)}\\
\cmidrule(l{0.5ex}r{0.5ex}){2-3}\cmidrule(l{0.5ex}r{0.5ex}){4-5}
    & Clean$\downarrow$ & Final$\downarrow$ & Fl-epe$\downarrow$ & Fl-all$\downarrow$ \\ 
\midrule
    PWC-Net~\citep{sun2018pwc} & 2.55 & 3.93 & 10.4 & 33.7\\
    RAFT~\citep{teed2020raft} & 1.43 & 2.71 & 5.04 & 17.4\\
    GMA~\citep{jiang2021learning} & 1.30 & 2.74 & 4.69 & 17.1\\
    SKFlow~\citep{sun2022skflow} & 1.22 & 2.46 & 4.27 & 15.5\\
    DIP~\citep{zheng2022dip} & 1.30 & 2.82 & 4.29 & 13.7\\
    EMD-L~\citep{deng2023explicit} & 0.88 & 2.55 & 4.12 & 13.5\\
    CRAFT~\citep{sui2022craft} & 1.27 & 2.79 & 4.88 & 17.5\\
    RPKNet~\citep{morimitsu2024recurrent} & 1.12 & 2.45 & - & 13.0\\
    GMFlowNet~\citep{zhao2022global} & 1.14 & 2.71 & 4.24 & 15.4\\
    FlowFormer~\citep{huang2022flowformer} & 1.01 & 2.40 & 4.09 & 14.7\\
    Flowformer++~\citep{shi2023flowformer++} & 0.90 & 2.30 & 3.93 & 14.2\\ 
    CCMR+~\citep{jahedi2024ccmr} & 0.98 & 2.36 & - & 12.9\\
    % AnyFlow~\citep{jung2023anyflow} & 
    MatchFlow(G)~\citep{dong2023rethinking} &1.03& 2.45 & 4.08 & 15.6 \\
    SEA-RAFT(L)~\citep{wang2024sea} &1.19 & 4.11 & 3.62 & 12.9\\
    AnyFlow~\citep{jung2023anyflow} & 1.10 & 2.52 & 3.76 & 12.4 \\
    SAMFlow~\citep{zhou2024samflow} & 0.87 & \textBF{2.11} & 3.44 & 12.3 \\
    FlowDiffuser~\citep{luo2024flowdiffuser} & \textBF{0.86} & 2.19 & 3.61 & 11.8 \\
    DPFlow~\citep{morimitsu2025dpflow} & 1.02 & 2.26 & 3.37 & 11.1 \\
    \midrule
    FlowNet2~\citep{ilg2017flownet} & 2.02 & 3.14 & 10.1 & 30.4 \\
    \textBF{WAFT-DAv2-a1} & 1.00 & 2.15 & \textBF{3.10} & \textBF{10.3} \\
    \textBF{WAFT-Twins-a2} & 1.02 & 2.46 & \textBF{2.98} & \textBF{9.9} \\
    \textBF{WAFT-DAv2-a2} & 1.01 & 2.49 & \textBF{3.28} & \textBF{10.9} \\
    \textBF{WAFT-DINOv3-a2} & 1.28 & 2.56 & 3.49 & 12.9 \\
\bottomrule
\end{tabular}
}
\caption{WAFT achieves the best cross-dataset generalization on KITTI(train), reducing the error by 11\%. We highlight all SOTA performance.}
\label{tab:zero-shot}
\end{wraptable}

Following previous work~\citep{teed2020raft, huang2022flowformer, sun2018pwc}, we train our model on FlyingChairs~\citep{dosovitskiy2015flownet} and  FlyingThings~\citep{mayer2016large}. Then we evaluate the performance on the training split of Sintel~\citep{butler2012naturalistic} and KITTI~\citep{geiger2013vision}. 

\myparagraph{Analysis} Results are shown in Table~\ref{tab:zero-shot}. WAFT achieves strong cross-dataset generalization. On KITTI (train), WAFT outperforms other methods by a large margin with an ImageNet-pretrained Twins backbone: It improves the endpoint-error (EPE) from 3.37 to 2.98 and Fl from 11.1 to 9.9. On Sintel (train), WAFT achieves performance close to state-of-the-art methods. Compared to the previous warping-based method~\citep{ilg2017flownet}, WAFT improves the performance by at least 31\%.

\subsection{Ablation Study}

\label{sec:ablation}

We conduct zero-shot ablations in Table~\ref{tab:ablations} on the training split of Sintel~\citep{butler2012naturalistic} and the sub-val split~\citep{wang2024sea} of Spring~\citep{mehl2023spring} based on WAFT-DAv2-a1. In all experiments, the models are trained on FlythingThings~\citep{mayer2016large} for 50k steps, with a batch size of 32 and learning rate $10^{-4}$. The average EPE and 1px are reported.

\myparagraph{\textcolor{CadetBlue}{Different Input Encoder}} Both pre-trained weights and adaptations are important to the performance. Note that the strong performance of WAFT is not merely from advanced backbones. Using a Twins backbone only pre-trained on ImageNet as adopted in Flowformer~\citep{huang2022flowformer}, WAFT ranks first on Spring, second on KITTI, and is competitive on Sintel. More details can be found in Table~\ref{tab:Sintel&KITTI},~\ref{tab:Spring}, and~\ref{tab:zero-shot}. 

\myparagraph{\textcolor{orange}{Different Recurrent Update Module}} The vision transformer design is crucial to iterative warping. We observe a significant performance drop when replacing the DPT-based recurrent update module with CNNs, highlighting the importance of modeling long-range dependence. This finding may help explain why early deep learning approaches~\citep{ilg2017flownet, ranjan2017optical} that implemented warping using CNNs underperformed compared to cost-volume-based methods~\citep{sun2018pwc, teed2020raft, huang2022flowformer}.

\myparagraph{\textcolor{Mahogany}{High-Resolution Indexing}} High-resolution indexing into feature maps using current flow estimates remarkably improves performance. We implement variants that index at 1/8 resolution by changing the patch size of DPT to $2\times 2$, and find that high-resolution indexing significantly improves 1px-outlier rate on Spring(sub-val)~\citep{mehl2023spring}. 

We also design a cost-volume-based variant following the common setup~\citep{wang2024sea, teed2020raft, huang2022flowformer} with look-up radius $4$ at 1/8 resolution, and find that it performs similarly to the warping counterpart (shown in Table~\ref{tab:ablations}) but costs $2.2\times$ training memory (21.2 GiB vs. 9.5 GiB).

\setlength\tabcolsep{.5em}
\begin{table*}[t]
    \centering
    \resizebox{1.0\linewidth}{!}{
    \begin{tabular}{lccccccccc}
    \toprule
        \multirow{2}{*}[\multirowcenter]{Experiment} & \multirow{2}{*}[\multirowcenter]{Input Enc.} & \multirow{2}{*}[\multirowcenter]{Rec. Upd.} & \multirow{2}{*}[\multirowcenter]{\#Steps} & \multirow{2}{*}[\multirowcenter]{\shortstack[c]{Index\\Reso.}} & \multicolumn{2}{c}{Sintel(train)} & \multicolumn{2}{c}{Spring(sub-val)} &\multirow{2}{*}[\multirowcenter]{\#MACs}\\
    \cmidrule(l{0.5ex}r{0.5ex}){6-7}\cmidrule(l{0.5ex}r{0.5ex}){8-9}
        & & & & & Clean$\downarrow$ & Final$\downarrow$ & EPE$\downarrow$ & 1px$\downarrow$ &\\ 
    \midrule
        WAFT-DAv2-a1 & DAv2-S+Res18 & DPT-S & 5 & 1/2 &1.18 & 2.33 & 0.27 & 1.43 & 858G \\
    \midrule
        \multirow{2}{*}{\textcolor{CadetBlue}{Different input enc.}} &  Res18 & \multirow{2}{*}{DPT-S}& \multirow{2}{*}{5} & \multirow{2}{*}{1/2} & 1.27 & 2.81 & 0.27 & 1.59 & 600G \\
        & DAv2-S & & & & 1.55 & 2.64 & 0.37 & 2.70 & 670G \\   
    \midrule
        \textcolor{CadetBlue}{DAv2 w/o pre-train} & DAv2-S+Res18 & DPT-S & 5 & 1/2 & 1.42 & 2.74 & 0.28 & 1.77 & 858G \\
    \midrule
        \multirow{2}{*}{\textcolor{orange}{Different rec. upd.}} & \multirow{2}{*}{DAv2-S+Res18} & Res18 & \multirow{2}{*}{5} & \multirow{2}{*}{1/2} & 7.23 & 6.84 & 0.45 & 2.93 & 1098G \\
        & & ConvGRU & & & 2.79 & 4.80 & 0.39 & 2.71 & 800G \\
    \midrule
        \textcolor{Mahogany}{1/8 reso. + warp} & DAv2-S+Res18 & DPT-S & 5 & 1/8 & 1.15 & 2.31 & 0.32 & 1.82 & 859G \\
    \midrule
        \textcolor{Mahogany}{1/8 reso. + corr.} & DAv2-S+Res18 & DPT-S & 5 & 1/8 & 1.10 & 2.45 & 0.33 & 1.74 & 883G \\
    \midrule
        \multirow{2}{*}{\textcolor{Plum}{Direct variants}} & \multirow{2}{*}{DAv2-B+Res18} & DPT-S & \multirow{2}{*}{1} & \multirow{2}{*}{1/2} & 2.36 & 3.43 & 0.59 & 10.5 & 1009G \\
        & & DPT-B & & & 2.37 & 3.38 & 0.61 & 11.1 & 1277G \\
    \midrule
        \textcolor{Blue}{Image-space warp} & DAv2-S+Res18 & DPT-S & 5 & 1/2 & 1.28 & 2.50 & 0.27 & 1.37 & 1902G \\
    \midrule
        \textcolor{Sepia}{Refine w/o warp} & DAv2-S+Res18 & DPT-S & 5 & 1/2 & 2.04 & 3.42 & 0.58 & 9.44 & 858G \\
    \midrule
        \textcolor{ForestGreen}{w/ Context} & DAv2-S+Res18 & DPT-S & 5 & 1/2 & 1.22 & 2.32 & 0.29 & 1.70 & 1005G \\
    \bottomrule
    \end{tabular}
    }
    \caption{We report the zero-shot ablation results on Sintel(train)~\citep{butler2012naturalistic} and Spring(sub-val)~\citep{mehl2023spring, wang2024sea}. The effect of changes can be identified through comparisons with the first row. See Section~\ref{sec:ablation} for details.}
    \label{tab:ablations}
    \vspace{-5pt}
\end{table*}

\myparagraph{\textcolor{Plum}{Direct vs. Iterative}} Iterative updates achieve better performance than direct regression within our meta-architecture. We implement direct regression by setting the number of iterations $T=1$. For fair comparison, we scale up the networks to match the computational cost of 5-iteration WAFT. Our results show that WAFT significantly outperforms these direct regression variants, indicating the effectiveness and high efficiency of the iterative paradigm. This finding aligns with the observation that existing direct methods either underperform~\citep{xu2022gmflow, weinzaepfel2022croco} or require substantially more computational cost~\citep{saxena2024surprising, weinzaepfel2022croco, weinzaepfel2023croco} compared to iterative approaches~\citep{teed2020raft, wang2024sea, huang2022flowformer, morimitsu2025dpflow}.

\myparagraph{\textcolor{Blue}{Warping Features vs. Pixels}} Feature-space warping is more effective than image-space warping, which is commonly used in classic methods~\citep{ma2022multiview, brox2004high, black1996robust} and early deep learning methods~\citep{ilg2017flownet, ranjan2017optical}. Feature-space warping does not need to re-extract features for the warped image in each iteration, significantly saving computational cost while achieving slightly better accuracy.

\myparagraph{\textcolor{Sepia}{Effectiveness of Warping}} It is possible to perform iterative updates without warping the features using the current flow estimates. We can simply use the original feature maps as input to the update module. Compared to this baseline, warping has significantly lower error with a negligible cost. This observation aligns with the conclusions of previous work~\citep{brox2004high}, which theoretically justifies the combination of warping and recurrent updates by framing it as a fixed-point iteration algorithm.

\myparagraph{\textcolor{ForestGreen}{Context Encoder}} Prior work~\citep{sun2018pwc, teed2020raft, wang2024sea, huang2022flowformer} has often used a context encoder that provides an extra input to the update module. Our ablation show that the context encoder is not necessary. The context encoder introduces additional computation overhead, but does not significantly affect performance. Previous work~\citep{wang2024sea} also points out that the context encoder can be regarded as a direct flow regressor, which functions similarly to the first iteration of WAFT.

\subsubsection*{Acknowledgments}
This work was partially supported by the National Science Foundation. 

\bibliography{iclr2026_conference}

@String(AAAI  = {AAAI})

@article{black1996robust,
  title={The robust estimation of multiple motions: Parametric and piecewise-smooth flow fields},
  author={Black, Michael J and Anandan, Paul},
  journal={Computer vision and image understanding},
  volume={63},
  number={1},
  pages={75--104},
  year={1996},
  publisher={Elsevier}
}

@article{simeoni2025dinov3,
  title={Dinov3},
  author={Sim{\'e}oni, Oriane and Vo, Huy V and Seitzer, Maximilian and Baldassarre, Federico and Oquab, Maxime and Jose, Cijo and Khalidov, Vasil and Szafraniec, Marc and Yi, Seungeun and Ramamonjisoa, Micha{\"e}l and others},
  journal={arXiv preprint arXiv:2508.10104},
  year={2025}
}

@article{janai2020computer,
  title={Computer vision for autonomous vehicles: Problems, datasets and state of the art},
  author={Janai, Joel and G{\"u}ney, Fatma and Behl, Aseem and Geiger, Andreas and others},
  journal={Foundations and trends{\textregistered} in computer graphics and vision},
  volume={12},
  number={1--3},
  pages={1--308},
  year={2020},
  publisher={Now Publishers, Inc.}
}

@inproceedings{memin1998multigrid,
  title={A multigrid approach for hierarchical motion estimation},
  author={Memin, Etienne and Perez, Patrick},
  booktitle={Sixth International Conference on Computer Vision (IEEE Cat. No. 98CH36271)},
  pages={933--938},
  year={1998},
  organization={IEEE}
}

@article{yang2024depthv2,
  title={Depth anything v2},
  author={Yang, Lihe and Kang, Bingyi and Huang, Zilong and Zhao, Zhen and Xu, Xiaogang and Feng, Jiashi and Zhao, Hengshuang},
  journal={Advances in Neural Information Processing Systems},
  volume={37},
  pages={21875--21911},
  year={2024}
}

@inproceedings{ranftl2021vision,
  title={Vision transformers for dense prediction},
  author={Ranftl, Ren{\'e} and Bochkovskiy, Alexey and Koltun, Vladlen},
  booktitle={Proceedings of the IEEE/CVF international conference on computer vision},
  pages={12179--12188},
  year={2021}
}

@inproceedings{rombach2022high,
  title={High-resolution image synthesis with latent diffusion models},
  author={Rombach, Robin and Blattmann, Andreas and Lorenz, Dominik and Esser, Patrick and Ommer, Bj{\"o}rn},
  booktitle={Proceedings of the IEEE/CVF conference on computer vision and pattern recognition},
  pages={10684--10695},
  year={2022}
}

@inproceedings{he2022masked,
  title={Masked autoencoders are scalable vision learners},
  author={He, Kaiming and Chen, Xinlei and Xie, Saining and Li, Yanghao and Doll{\'a}r, Piotr and Girshick, Ross},
  booktitle={Proceedings of the IEEE/CVF conference on computer vision and pattern recognition},
  pages={16000--16009},
  year={2022}
}

@inproceedings{kirillov2023segment,
  title={Segment anything},
  author={Kirillov, Alexander and Mintun, Eric and Ravi, Nikhila and Mao, Hanzi and Rolland, Chloe and Gustafson, Laura and Xiao, Tete and Whitehead, Spencer and Berg, Alexander C and Lo, Wan-Yen and others},
  booktitle={Proceedings of the IEEE/CVF international conference on computer vision},
  pages={4015--4026},
  year={2023}
}

@article{oquab2023dinov2,
  title={Dinov2: Learning robust visual features without supervision},
  author={Oquab, Maxime and Darcet, Timoth{\'e}e and Moutakanni, Th{\'e}o and Vo, Huy and Szafraniec, Marc and Khalidov, Vasil and Fernandez, Pierre and Haziza, Daniel and Massa, Francisco and El-Nouby, Alaaeldin and others},
  journal={arXiv preprint arXiv:2304.07193},
  year={2023}
}

@inproceedings{zhao2024hybrid,
  title={Hybrid Cost Volume for Memory-Efficient Optical Flow},
  author={Zhao, Yang and Xu, Gangwei and Wu, Gang},
  booktitle={Proceedings of the 32nd ACM International Conference on Multimedia},
  pages={8740--8749},
  year={2024}
}

@article{xu2023memory,
  title={Memory-efficient optical flow via radius-distribution orthogonal cost volume},
  author={Xu, Gangwei and Chen, Shujun and Jia, Hao and Feng, Miaojie and Yang, Xin},
  journal={arXiv preprint arXiv:2312.03790},
  year={2023}
}

@inproceedings{brox2004high,
  title={High accuracy optical flow estimation based on a theory for warping},
  author={Brox, Thomas and Bruhn, Andr{\'e}s and Papenberg, Nils and Weickert, Joachim},
  booktitle={Computer Vision-ECCV 2004: 8th European Conference on Computer Vision, Prague, Czech Republic, May 11-14, 2004. Proceedings, Part IV 8},
  pages={25--36},
  year={2004},
  organization={Springer}
}

@article{morimitsu2025dpflow,
  title={DPFlow: Adaptive Optical Flow Estimation with a Dual-Pyramid Framework},
  author={Morimitsu, Henrique and Zhu, Xiaobin and Cesar Jr, Roberto M and Ji, Xiangyang and Yin, Xu-Cheng},
  journal={arXiv preprint arXiv:2503.14880},
  year={2025}
}

@inproceedings{luo2024flowdiffuser,
  title={Flowdiffuser: Advancing optical flow estimation with diffusion models},
  author={Luo, Ao and Li, Xin and Yang, Fan and Liu, Jiangyu and Fan, Haoqiang and Liu, Shuaicheng},
  booktitle={Proceedings of the IEEE/CVF Conference on Computer Vision and Pattern Recognition},
  pages={19167--19176},
  year={2024}
}

@inproceedings{wang2024sea,
  title={Sea-raft: Simple, efficient, accurate raft for optical flow},
  author={Wang, Yihan and Lipson, Lahav and Deng, Jia},
  booktitle={European Conference on Computer Vision},
  pages={36--54},
  year={2024},
  organization={Springer}
}

@inproceedings{zhou2024samflow,
  title={Samflow: Eliminating any fragmentation in optical flow with segment anything model},
  author={Zhou, Shili and He, Ruian and Tan, Weimin and Yan, Bo},
  booktitle={Proceedings of the AAAI Conference on Artificial Intelligence},
  volume={38},
  number={7},
  pages={7695--7703},
  year={2024}
}

@inproceedings{he2016deep,
  title={Deep residual learning for image recognition},
  author={He, Kaiming and Zhang, Xiangyu and Ren, Shaoqing and Sun, Jian},
  booktitle={Proceedings of the IEEE conference on computer vision and pattern recognition},
  pages={770--778},
  year={2016}
}

@inproceedings{xu2017accurate,
  title={Accurate optical flow via direct cost volume processing},
  author={Xu, Jia and Ranftl, Ren{\'e} and Koltun, Vladlen},
  booktitle={Proceedings of the IEEE Conference on Computer Vision and Pattern Recognition},
  pages={1289--1297},
  year={2017}
}

@inproceedings{luo2023gaflow,
  title={Gaflow: Incorporating gaussian attention into optical flow},
  author={Luo, Ao and Yang, Fan and Li, Xin and Nie, Lang and Lin, Chunyu and Fan, Haoqiang and Liu, Shuaicheng},
  booktitle={Proceedings of the IEEE/CVF International Conference on Computer Vision},
  pages={9642--9651},
  year={2023}
}

@article{jahedi2023ms,
  title={MS-RAFT+: High Resolution Multi-Scale RAFT},
  author={Jahedi, Azin and Luz, Maximilian and Rivinius, Marc and Mehl, Lukas and Bruhn, Andr{\'e}s},
  journal={International Journal of Computer Vision},
  pages={1--22},
  year={2023},
  publisher={Springer}
}

@article{saxena2024surprising,
  title={The surprising effectiveness of diffusion models for optical flow and monocular depth estimation},
  author={Saxena, Saurabh and Herrmann, Charles and Hur, Junhwa and Kar, Abhishek and Norouzi, Mohammad and Sun, Deqing and Fleet, David J},
  journal={Advances in Neural Information Processing Systems},
  volume={36},
  year={2024}
}

@inproceedings{richter2017playing,
  title={Playing for benchmarks},
  author={Richter, Stephan R and Hayder, Zeeshan and Koltun, Vladlen},
  booktitle={Proceedings of the IEEE International Conference on Computer Vision},
  pages={2213--2222},
  year={2017}
}

@article{weinzaepfel2022croco,
  title={CroCo: Self-Supervised Pre-training for 3D Vision Tasks by Cross-View Completion},
  author={Weinzaepfel, Philippe and Leroy, Vincent and Lucas, Thomas and Br{\'e}gier, Romain and Cabon, Yohann and Arora, Vaibhav and Antsfeld, Leonid and Chidlovskii, Boris and Csurka, Gabriela and Revaud, J{\'e}r{\^o}me},
  journal={Advances in Neural Information Processing Systems},
  volume={35},
  pages={3502--3516},
  year={2022}
}

@inproceedings{deng2023explicit,
  title={Explicit motion disentangling for efficient optical flow estimation},
  author={Deng, Changxing and Luo, Ao and Huang, Haibin and Ma, Shaodan and Liu, Jiangyu and Liu, Shuaicheng},
  booktitle={Proceedings of the IEEE/CVF International Conference on Computer Vision},
  pages={9521--9530},
  year={2023}
}

@inproceedings{zach2007duality,
  title={A duality based approach for realtime tv-l 1 optical flow},
  author={Zach, Christopher and Pock, Thomas and Bischof, Horst},
  booktitle={Pattern Recognition: 29th DAGM Symposium, Heidelberg, Germany, September 12-14, 2007. Proceedings 29},
  pages={214--223},
  year={2007},
  organization={Springer}
}

@inproceedings{chen2016full,
  title={Full flow: Optical flow estimation by global optimization over regular grids},
  author={Chen, Qifeng and Koltun, Vladlen},
  booktitle={Proceedings of the IEEE conference on computer vision and pattern recognition},
  pages={4706--4714},
  year={2016}
}

@inproceedings{wang2020tartanair,
  title={Tartanair: A dataset to push the limits of visual slam},
  author={Wang, Wenshan and Zhu, Delong and Wang, Xiangwei and Hu, Yaoyu and Qiu, Yuheng and Wang, Chen and Hu, Yafei and Kapoor, Ashish and Scherer, Sebastian},
  booktitle={2020 IEEE/RSJ International Conference on Intelligent Robots and Systems (IROS)},
  pages={4909--4916},
  year={2020},
  organization={IEEE}
}

@inproceedings{luo2022kpa,
  title={Learning Optical Flow With Kernel Patch Attention},
  author={Luo, Ao and Yang, Fan and Li, Xin and Liu, Shuaicheng},
  booktitle={Proceedings of the IEEE/CVF Conference on Computer Vision and Pattern Recognition},
  pages={8906--8915},
  year={2022}
}

@inproceedings{huang2022flowformer,
  title={Flowformer: A transformer architecture for optical flow},
  author={Huang, Zhaoyang and Shi, Xiaoyu and Zhang, Chao and Wang, Qiang and Cheung, Ka Chun and Qin, Hongwei and Dai, Jifeng and Li, Hongsheng},
  booktitle={European Conference on Computer Vision},
  pages={668--685},
  year={2022},
  organization={Springer}
}

@inproceedings{sui2022craft,
  title={Craft: Cross-attentional flow transformer for robust optical flow},
  author={Sui, Xiuchao and Li, Shaohua and Geng, Xue and Wu, Yan and Xu, Xinxing and Liu, Yong and Goh, Rick and Zhu, Hongyuan},
  booktitle={Proceedings of the IEEE/CVF conference on Computer Vision and Pattern Recognition},
  pages={17602--17611},
  year={2022}
}

@inproceedings{zhao2022global,
  title={Global matching with overlapping attention for optical flow estimation},
  author={Zhao, Shiyu and Zhao, Long and Zhang, Zhixing and Zhou, Enyu and Metaxas, Dimitris},
  booktitle={Proceedings of the IEEE/CVF Conference on Computer Vision and Pattern Recognition},
  pages={17592--17601},
  year={2022}
}

@article{sun2022skflow,
  title={Skflow: Learning optical flow with super kernels},
  author={Sun, Shangkun and Chen, Yuanqi and Zhu, Yu and Guo, Guodong and Li, Ge},
  journal={Advances in Neural Information Processing Systems},
  volume={35},
  pages={11313--11326},
  year={2022}
}

@inproceedings{xu2022gmflow,
  title={Gmflow: Learning optical flow via global matching},
  author={Xu, Haofei and Zhang, Jing and Cai, Jianfei and Rezatofighi, Hamid and Tao, Dacheng},
  booktitle={Proceedings of the IEEE/CVF conference on computer vision and pattern recognition},
  pages={8121--8130},
  year={2022}
}

@inproceedings{shi2023flowformer++,
  title={Flowformer++: Masked cost volume autoencoding for pretraining optical flow estimation},
  author={Shi, Xiaoyu and Huang, Zhaoyang and Li, Dasong and Zhang, Manyuan and Cheung, Ka Chun and See, Simon and Qin, Hongwei and Dai, Jifeng and Li, Hongsheng},
  booktitle={Proceedings of the IEEE/CVF Conference on Computer Vision and Pattern Recognition},
  pages={1599--1610},
  year={2023}
}

@article{xu2023unifying,
  title={Unifying flow, stereo and depth estimation},
  author={Xu, Haofei and Zhang, Jing and Cai, Jianfei and Rezatofighi, Hamid and Yu, Fisher and Tao, Dacheng and Geiger, Andreas},
  journal={IEEE Transactions on Pattern Analysis and Machine Intelligence},
  year={2023},
  publisher={IEEE}
}

@inproceedings{weinzaepfel2023croco,
  title={CroCo v2: Improved Cross-view Completion Pre-training for Stereo Matching and Optical Flow},
  author={Weinzaepfel, Philippe and Lucas, Thomas and Leroy, Vincent and Cabon, Yohann and Arora, Vaibhav and Br{\'e}gier, Romain and Csurka, Gabriela and Antsfeld, Leonid and Chidlovskii, Boris and Revaud, J{\'e}r{\^o}me},
  booktitle={Proceedings of the IEEE/CVF International Conference on Computer Vision},
  pages={17969--17980},
  year={2023}
}

@inproceedings{dong2023rethinking,
  title={Rethinking Optical Flow from Geometric Matching Consistent Perspective},
  author={Dong, Qiaole and Cao, Chenjie and Fu, Yanwei},
  booktitle={Proceedings of the IEEE/CVF Conference on Computer Vision and Pattern Recognition},
  pages={1337--1347},
  year={2023}
}

@inproceedings{zheng2022dip,
  title={Dip: Deep inverse patchmatch for high-resolution optical flow},
  author={Zheng, Zihua and Nie, Ni and Ling, Zhi and Xiong, Pengfei and Liu, Jiangyu and Wang, Hao and Li, Jiankun},
  booktitle={Proceedings of the IEEE/CVF Conference on Computer Vision and Pattern Recognition},
  pages={8925--8934},
  year={2022}
}

@inproceedings{sun2018optical,
  title={Optical flow guided feature: A fast and robust motion representation for video action recognition},
  author={Sun, Shuyang and Kuang, Zhanghui and Sheng, Lu and Ouyang, Wanli and Zhang, Wei},
  booktitle={Proceedings of the IEEE conference on computer vision and pattern recognition},
  pages={1390--1399},
  year={2018}
}

@inproceedings{piergiovanni2019representation,
  title={Representation flow for action recognition},
  author={Piergiovanni, AJ and Ryoo, Michael S},
  booktitle={Proceedings of the IEEE/CVF Conference on Computer Vision and Pattern Recognition},
  pages={9945--9953},
  year={2019}
}

@article{xu2019quadratic,
  title={Quadratic video interpolation},
  author={Xu, Xiangyu and Siyao, Li and Sun, Wenxiu and Yin, Qian and Yang, Ming-Hsuan},
  journal={Advances in Neural Information Processing Systems},
  volume={32},
  year={2019}
}

@article{huang2020rife,
  title={Rife: Real-time intermediate flow estimation for video frame interpolation},
  author={Huang, Zhewei and Zhang, Tianyuan and Heng, Wen and Shi, Boxin and Zhou, Shuchang},
  journal={arXiv preprint arXiv:2011.06294},
  year={2020}
}

@article{horn1981determining,
  title={Determining optical flow},
  author={Horn, Berthold KP and Schunck, Brian G},
  journal={Artificial intelligence},
  volume={17},
  number={1-3},
  pages={185--203},
  year={1981},
  publisher={Elsevier}
}

@inproceedings{dosovitskiy2015flownet,
  title={Flownet: Learning optical flow with convolutional networks},
  author={Dosovitskiy, Alexey and Fischer, Philipp and Ilg, Eddy and Hausser, Philip and Hazirbas, Caner and Golkov, Vladimir and Van Der Smagt, Patrick and Cremers, Daniel and Brox, Thomas},
  booktitle={Proceedings of the IEEE international conference on computer vision},
  pages={2758--2766},
  year={2015}
}

@inproceedings{ilg2017flownet,
  title={Flownet 2.0: Evolution of optical flow estimation with deep networks},
  author={Ilg, Eddy and Mayer, Nikolaus and Saikia, Tonmoy and Keuper, Margret and Dosovitskiy, Alexey and Brox, Thomas},
  booktitle={Proceedings of the IEEE conference on computer vision and pattern recognition},
  pages={2462--2470},
  year={2017}
}

@inproceedings{ranjan2017optical,
  title={Optical flow estimation using a spatial pyramid network},
  author={Ranjan, Anurag and Black, Michael J},
  booktitle={Proceedings of the IEEE conference on computer vision and pattern recognition},
  pages={4161--4170},
  year={2017}
}

@inproceedings{sun2018pwc,
  title={Pwc-net: Cnns for optical flow using pyramid, warping, and cost volume},
  author={Sun, Deqing and Yang, Xiaodong and Liu, Ming-Yu and Kautz, Jan},
  booktitle={Proceedings of the IEEE conference on computer vision and pattern recognition},
  pages={8934--8943},
  year={2018}
}

@inproceedings{hui2018liteflownet,
  title={Liteflownet: A lightweight convolutional neural network for optical flow estimation},
  author={Hui, Tak-Wai and Tang, Xiaoou and Loy, Chen Change},
  booktitle={Proceedings of the IEEE conference on computer vision and pattern recognition},
  pages={8981--8989},
  year={2018}
}

@inproceedings{teed2020raft,
  title={Raft: Recurrent all-pairs field transforms for optical flow},
  author={Teed, Zachary and Deng, Jia},
  booktitle={European conference on computer vision},
  pages={402--419},
  year={2020},
  organization={Springer}
}

@article{jiang2021learning,
  title={Learning to Estimate Hidden Motions with Global Motion Aggregation},
  author={Jiang, Shihao and Campbell, Dylan and Lu, Yao and Li, Hongdong and Hartley, Richard},
  journal={arXiv preprint arXiv:2104.02409},
  year={2021}
}

@article{dosovitskiy2020image,
  title={An image is worth 16x16 words: Transformers for image recognition at scale},
  author={Dosovitskiy, Alexey and Beyer, Lucas and Kolesnikov, Alexander and Weissenborn, Dirk and Zhai, Xiaohua and Unterthiner, Thomas and Dehghani, Mostafa and Minderer, Matthias and Heigold, Georg and Gelly, Sylvain and others},
  journal={arXiv preprint arXiv:2010.11929},
  year={2020}
}

@inproceedings{menze2015object,
  title={Object scene flow for autonomous vehicles},
  author={Menze, Moritz and Geiger, Andreas},
  booktitle={Proceedings of the IEEE conference on computer vision and pattern recognition},
  pages={3061--3070},
  year={2015}
}

@inproceedings{jahedi2024ccmr,
  title={CCMR: High Resolution Optical Flow Estimation via Coarse-to-Fine Context-Guided Motion Reasoning},
  author={Jahedi, Azin and Luz, Maximilian and Rivinius, Marc and Bruhn, Andr{\'e}s},
  booktitle={Proceedings of the IEEE/CVF Winter Conference on Applications of Computer Vision},
  pages={6899--6908},
  year={2024}
}

@article{chu2021twins,
  title={Twins: Revisiting spatial attention design in vision transformers},
  author={Chu, Xiangxiang and Tian, Zhi and Wang, Yuqing and Zhang, Bo and Ren, Haibing and Wei, Xiaolin and Xia, Huaxia and Shen, Chunhua},
  journal={arXiv preprint arXiv:2104.13840},
  year={2021}
}

@inproceedings{butler2012naturalistic,
  title={A naturalistic open source movie for optical flow evaluation},
  author={Butler, Daniel J and Wulff, Jonas and Stanley, Garrett B and Black, Michael J},
  booktitle={European conference on computer vision},
  pages={611--625},
  year={2012},
  organization={Springer}
}

@article{geiger2013vision,
  title={Vision meets robotics: The kitti dataset},
  author={Geiger, Andreas and Lenz, Philip and Stiller, Christoph and Urtasun, Raquel},
  journal={The International Journal of Robotics Research},
  volume={32},
  number={11},
  pages={1231--1237},
  year={2013},
  publisher={Sage Publications Sage UK: London, England}
}

@inproceedings{jung2023anyflow,
  title={AnyFlow: Arbitrary Scale Optical Flow with Implicit Neural Representation},
  author={Jung, Hyunyoung and Hui, Zhuo and Luo, Lei and Yang, Haitao and Liu, Feng and Yoo, Sungjoo and Ranjan, Rakesh and Demandolx, Denis},
  booktitle={Proceedings of the IEEE/CVF Conference on Computer Vision and Pattern Recognition},
  pages={5455--5465},
  year={2023}
}

@inproceedings{ma2022multiview,
  title={Multiview stereo with cascaded epipolar raft},
  author={Ma, Zeyu and Teed, Zachary and Deng, Jia},
  booktitle={European Conference on Computer Vision},
  pages={734--750},
  year={2022},
  organization={Springer}
}

@article{zuo2022view,
  title={View synthesis with sculpted neural points},
  author={Zuo, Yiming and Deng, Jia},
  journal={arXiv preprint arXiv:2205.05869},
  year={2022}
}

@inproceedings{mayer2016large,
  title={A large dataset to train convolutional networks for disparity, optical flow, and scene flow estimation},
  author={Mayer, Nikolaus and Ilg, Eddy and Hausser, Philip and Fischer, Philipp and Cremers, Daniel and Dosovitskiy, Alexey and Brox, Thomas},
  booktitle={Proceedings of the IEEE conference on computer vision and pattern recognition},
  pages={4040--4048},
  year={2016}
}

@inproceedings{kondermann2016hci,
  title={The hci benchmark suite: Stereo and flow ground truth with uncertainties for urban autonomous driving},
  author={Kondermann, Daniel and Nair, Rahul and Honauer, Katrin and Krispin, Karsten and Andrulis, Jonas and Brock, Alexander and Gussefeld, Burkhard and Rahimimoghaddam, Mohsen and Hofmann, Sabine and Brenner, Claus and others},
  booktitle={Proceedings of the IEEE Conference on Computer Vision and Pattern Recognition Workshops},
  pages={19--28},
  year={2016}
}

@article{sun2019models,
  title={Models matter, so does training: An empirical study of cnns for optical flow estimation},
  author={Sun, Deqing and Yang, Xiaodong and Liu, Ming-Yu and Kautz, Jan},
  journal={IEEE transactions on pattern analysis and machine intelligence},
  volume={42},
  number={6},
  pages={1408--1423},
  year={2019},
  publisher={IEEE}
}

@article{leroy2023win,
  title={Win-Win: Training High-Resolution Vision Transformers from Two Windows},
  author={Leroy, Vincent and Revaud, Jerome and Lucas, Thomas and Weinzaepfel, Philippe},
  journal={arXiv preprint arXiv:2310.00632},
  year={2023}
}

@inproceedings{zhao2020maskflownet,
  title={Maskflownet: Asymmetric feature matching with learnable occlusion mask},
  author={Zhao, Shengyu and Sheng, Yilun and Dong, Yue and Chang, Eric I and Xu, Yan and others},
  booktitle={Proceedings of the IEEE/CVF Conference on Computer Vision and Pattern Recognition},
  pages={6278--6287},
  year={2020}
}

@article{zhao2020improved,
  title={Improved two-stream model for human action recognition},
  author={Zhao, Yuxuan and Man, Ka Lok and Smith, Jeremy and Siddique, Kamran and Guan, Sheng-Uei},
  journal={EURASIP Journal on Image and Video Processing},
  volume={2020},
  number={1},
  pages={1--9},
  year={2020},
  publisher={Springer}
}

@article{liu2020video,
  title={Video frame interpolation via optical flow estimation with image inpainting},
  author={Liu, Xiaozhang and Liu, Hui and Lin, Yuxiu},
  journal={International Journal of Intelligent Systems},
  volume={35},
  number={12},
  pages={2087--2102},
  year={2020},
  publisher={Wiley Online Library}
}

@inproceedings{mehl2023spring,
  title={Spring: A High-Resolution High-Detail Dataset and Benchmark for Scene Flow, Optical Flow and Stereo},
  author={Mehl, Lukas and Schmalfuss, Jenny and Jahedi, Azin and Nalivayko, Yaroslava and Bruhn, Andr{\'e}s},
  booktitle={Proceedings of the IEEE/CVF Conference on Computer Vision and Pattern Recognition},
  pages={4981--4991},
  year={2023}
}

@article{morimitsu2024recurrent,
  title={Recurrent Partial Kernel Network for Efficient Optical Flow Estimation},
  author={Morimitsu, Henrique and Zhu, Xiaobin and Ji, Xiangyang and Yin, Xu-Cheng},
  year={2024}
}
\bibliographystyle{iclr2026_conference}

\appendix
\section{Appendix}
\label{sec:appendix}

\setlength\tabcolsep{.6em}
\begin{table*}
    \centering
    \resizebox{1.0\linewidth}{!}{
    \begin{tabular}{lcccccc}
    \toprule
        Sequence & WAFT-Twins-a2 & DPFlow & Flowformer++ & FlowDiffuser & DDVM & SAMFlow\\
    \midrule
        Perturbed Market 3 & 0.893 (0.460) & 0.892 (0.423) & 0.958 (0.511) & 0.897 (0.490) & 0.787 (0.372) & 0.932 (0.493)\\
        Perturbed Shaman 1 & 0.174 (0.163) & 0.213 (0.201) & 0.251 (0.236) & 0.267 (0.255) & 0.219 (0.196) & 0.241 (0.219)\\
        Ambush 1 & 20.561 (2.526) & 8.366 (2.970) & 6.610 (2.605) & 7.224 (2.583) & 29.33 (14.07) & 10.586 (2.733) \\
        Ambush 3 & 3.173 (2.048) & 3.019 (1.729) & 2.939 (1.816) & 3.148 (1.828) & 2.855 (3.016) & 3.411 (1.779)\\
        Bamboo 3 & 0.460 (0.423) & 0.486 (0.438) & 0.546 (0.513) & 0.594 (0.508) & 0.415 (0.380) & 0.522 (0.473)\\
        Cave 3 & 2.199 (1.567) & 2.341 (1.631) & 2.344 (1.477) & 2.464 (1.433) & 2.042 (1.658) & 2.475 (1.445)\\
        Market 1 & 0.851 (0.384) & 0.890 (0.491) & 1.073 (0.550) & 1.238 (0.517) & 0.719 (0.467) & 1.060 (0.491)\\
        Market 4 & 7.513 (3.933) & 7.939 (3.834) & 8.086 (4.450) & 8.024 (4.027) & 5.517 (3.971) & 6.636 (3.680)\\
        Mountain 2 & 0.366 (0.087) & 0.177 (0.078) & 0.288 (0.118) & 0.409 (0.101) & 0.176 (0.095) & 0.500 (0.217)\\
        Temple 1 & 0.511 (0.297) & 0.511 (0.302) & 0.657 (0.359) & 0.789 (0.340) & 0.452 (0.284) & 0.853 (0.340)\\
        Tiger &  0.463 (0.333) & 0.571 (0.411) & 0.595 (0.430) & 0.636 (0.391) & 0.413 (0.344) & 0.573 (0.392)\\
        Wall & 1.708 (0.979) & 2.011 (1.231) & 1.727 (0.793) & 1.692 (0.734) & 1.639 (3.210) & 2.105 (0.780)\\
        Avg & 2.393 (1.015) & 1.975 (1.046) & 1.943 (1.073) & 2.026 (1.016) & 2.475 (1.754) & 0.995 (2.080) \\
        Avg (w/o Ambush 1) & 1.639 (0.952) & 1.710 (0.966) & 1.750 (1.010) & 1.810 (0.951) & 1.360 (1.242) & 1.727 (0.923) \\
    \bottomrule
    \end{tabular}
    }
    \caption{We report the endpoint-error (EPE) on all sequences of Sintel~\citep{butler2012naturalistic}, shown in the format ``final-epe (clean-epe)''. We highlight the best result on each sequence.}
    \label{tab:sintel-seq}
\end{table*}

\myparagraph{Sintel Results} We show the sequence-wise results of several representative methods~\citep{morimitsu2025dpflow, huang2022flowformer, luo2024flowdiffuser, saxena2024surprising, zhou2024samflow} on Sintel in Table~\ref{tab:sintel-seq}. The sequence `Ambush 1' appears to be an outlier which severely affects the average EPE on the final split. Given the same ImageNet-pretrained Twins backbone, WAFT outperforms SOTA Flowformer++ when `Ambush 1' is excluded.

\end{document}